\documentclass{article}

     \usepackage[preprint]{neurips_2023}

\usepackage[utf8]{inputenc} % allow utf-8 input
\usepackage[T1]{fontenc}    % use 8-bit T1 fonts
\usepackage{hyperref}       % hyperlinks
\usepackage{url}            % simple URL typesetting
\usepackage{booktabs}       % professional-quality tables
\usepackage{amsfonts}       % blackboard math symbols
\usepackage{nicefrac}       % compact symbols for 1/2, etc.
\usepackage{microtype}      % microtypography
\usepackage{xcolor}         % colors
\usepackage{amsmath}

\usepackage{graphicx}

\usepackage{algorithm}
\usepackage{algpseudocode}
\usepackage{placeins}

\title{Do LLMs Know When to Flip a Coin? Strategic Randomization through Reasoning and Experience}

\author{%
  Lingyu Yang \\
  Shanghai Jiao Tong University\\
  \texttt{jlnhbyu.yang@sjtu.edu.cn} \\
}

\begin{document}

\maketitle

\begin{abstract}
  Strategic randomization is a key principle in game theory, yet it remains underexplored in large language models (LLMs). Prior work often conflates the cognitive decision to randomize with the mechanical generation of randomness, leading to incomplete evaluations. To address this, we propose a novel zero-sum game inspired by the Tian Ji Horse Race, where the Nash equilibrium corresponds to a maximal entropy strategy. The game's complexity masks this property from untrained humans and underdeveloped LLMs. We evaluate five LLMs across prompt styles—framed, neutral, and hinted—using competitive multi-tournament gameplay with system-provided random choices, isolating the decision to randomize. Results show that weaker models remain deterministic regardless of prompts, while stronger models exhibit increased randomization under explicit hints. When facing weaker models, strong LLMs adopt deterministic strategies to exploit biases, but converge toward equilibrium play when facing peers. Through win/loss outcomes and Bayes factor analysis, we demonstrate meaningful variation in LLMs’ strategic reasoning capabilities, highlighting opportunities for improvement in abstract reasoning and adaptive learning. We make our implementation publicly available at \url{https://github.com/ocelopus/llm-when-to-throw-coin} to ensure full reproducibility.
\end{abstract}

\section{Introduction}

 large language models (LLMs) have demonstrated remarkable capabilities across a wide range of domains, including strategic reasoning and decision-making in game-like scenarios. While much attention has been given to their ability to devise clever strategies or simulate human-like behavior, a critical question remains underexplored: can LLMs recognize when randomization is strategically optimal, and do they deliberately choose to randomize? This question lies at the intersection of game theory, cognitive reasoning, and machine behavior.

Randomization plays a pivotal role in many competitive settings. In games such as poker or rock-paper-scissors, adopting a randomized strategy can prevent opponents from predicting one’s moves and exploiting deterministic patterns. The concept of Nash equilibrium in zero-sum games often prescribes randomized (mixed) strategies as optimal solutions. However, identifying and implementing such strategies requires not only computational capability but also conceptual understanding of strategic depth and adaptability.

Prior studies evaluating LLMs in strategic environments either overlook the importance of randomization, or often conflate two distinct challenges: (1) the mechanical generation of randomness, which LLMs are known to perform poorly even with temperature adjustments, and (2) the cognitive decision to randomize, which involves recognizing that randomization is the optimal approach in a given context. This conflation leads to incomplete or misleading assessments of an LLM's strategic reasoning abilities.

To address this gap, we introduce a novel experimental framework centered around a structurally complex game inspired by the ancient Chinese tale of Tian Ji’s Horse Race. This game features a unique Nash equilibrium in which uniformly random selection among available actions is optimal. Crucially, we decouple the act of generating randomness from the decision to use it by providing LLMs with system-generated random choices. This allows us to isolate and evaluate whether LLMs can reason about the value of randomization in strategic contexts.

Our findings reveal meaningful differences among current LLMs in their ability to adopt randomized strategies, especially when prompted with explicit hints or engaged in repeated gameplay. We show that while stronger models can be guided toward optimal randomization, weaker models tend to remain deterministic regardless of prompting. These results provide insights into the current limitations and potential for improvement in LLMs’ strategic reasoning capabilities.

\section{Related Work}

\subsection{Game-Theoretic Evaluation of LLMs}

Recent efforts have explored the application of large language models (LLMs) in strategic and game-theoretic domains, with benchmarks such as \citet{NEURIPS2024_31911709} and \citet{tang2025dsgbenchdiversestrategicgame} assessing abilities like strategic thinking, adaptability, and multi-agent reasoning. However, these evaluations, albeit quite comprehensive, overlook randomization and often rely on heuristic or weighted-average metrics that lack rigorous theoretical grounding in game theory. Moreover, while some studies attempt to probe LLM behavior in games like rock-paper-scissors, such well-known and structurally simple games may not sufficiently challenge higher-order reasoning or expose nuanced strategic capabilities.

\subsection{LLMs and Randomness}

A growing body of research highlights the limitations of LLMs in generating truly random sequences. \citet{vidler2025playinggameslargelanguage} demonstrates that even with temperature adjustments, LLMs struggle to produce uniform randomness—an essential component in many game-theoretic strategies. However, their insistence on letting LLMs generate the random number limits the scope for whether LLMs can reason about when randomization is strategically optimal, rather than merely failing to generate random numbers.

\subsection{Crucial Difference in This Work}
Our work addresses this gap by decoupling the mechanism of randomness from the decision to randomize. Rather than asking LLMs to generate random choices themselves, we provide system-generated random options and observe whether and how models choose to use them. This cleanly isolates the cognitive aspect of strategic randomization — a critical step toward understanding whether LLMs can reason about optimal mixed strategies in non-trivial games.

\section{Methodology}

\subsection{Game Design}

We introduce a novel strategic game inspired by the ancient Chinese story of Tian Ji’s horse racing. The game is designed to test an LLM's ability to reason about and adopt randomized strategies when such behavior constitutes the Nash equilibrium.

Each session of the game is referred to as a \textbf{tournament}:

\begin{itemize}
    \item Each player begins with $ N $ horses, each having a unique speed from 1 (slowest) to $ N $ (fastest).
    \item A tournament consists of $ N $ rounds.
    \item In each round:
    \begin{itemize}
        \item Both players simultaneously select one unused horse.
        \item The horse(s) with the highest speed win a reward of 1.0 point, which is divided equally among all fastest horses in that round.
        \item Once used, a horse cannot be selected again.
    \end{itemize}
    \item At the end of the tournament, the player with the highest total reward wins.
\end{itemize}

This is a zero-sum, symmetric game with perfect information, and thus admits a Nash equilibrium in mixed strategies, where both players achieve an expected utility of zero. To identify the structure of this equilibrium, we numerically solve a Bellman equation under the assumption that the opponent plays according to a maximal entropy strategy—i.e., uniformly random over their remaining horses at each round. The result is that any counteractive policy yields an expected utility of 0, which confirms that uniform randomization constitutes a Nash equilibrium. Details of the numerical implementation and full results are provided in a separate supplementary document.

To demonstrate human learning dynamics, we conducted a small-scale human experiment. Initially, the opponent attempted deterministic strategies based on intuitive heuristics (e.g., trying to sacrifice slow horses to waste fast counterparts), but suffered consistent losses due to predictability. After several rounds, the opponent adopted randomization through extrinsic means of prime numbers, leading to balanced outcomes. This shows that even without formal game-theoretic knowledge or good intrinsic randomization mechanisms, humans can learn to randomize through experience.

\subsection{Evaluation Framework}

Our evaluation framework enables us to cleanly isolate the \textbf{decision to randomize} from the \textbf{mechanics of random number generation}, addressing a key limitation in prior work.

\paragraph{Interaction Loop}

The gameplay loop proceeds as follows:

\begin{enumerate}
    \item  The game state is presented to both LLMs.
    \item For each round, a system-generated random choice is provided to each model: this is a uniformly sampled horse from its remaining set.
    \item The LLM is prompted to make a move, possibly using or ignoring the suggested random choice.
    \item The chosen action is logged, and the game state is updated.
    \item At the end of the tournament, the winner is determined, and Bayes factor analysis is performed.
\end{enumerate}

This structure ensures that any observed randomness in the LLM’s actions stems from a deliberate \textbf{choice to randomize}, not from uncontrollable and biased noise in its generation process.

\begin{algorithm}[h!]
\caption{Evaluate LLMs' Strategic Randomization}
\label{alg:evaluation}
\textbf{Input: }A set of LLMs $\mathcal{M} = \{M_1, M_2, ..., M_5\}$, number of tournaments $K$, number of horses $N$.\\
\textbf{Output: }Win/loss matrix $W$ and average log Bayes factor matrix $B$ for all model pairs.
\begin{algorithmic}[1]

    \ForAll{pairs of models $(M_i, M_j) \in \mathcal{M} \times \mathcal{M}$}
      \State Prompt models with game background information
        \For{$k = 1$ to $K$}
            \State Initialize game state: each player has $N$ unused horses with speeds $1$ to $N$
            \While{game not finished}
                \State Provide both models with current game state
                \State Sample a random horse uniformly from remaining ones for each model
                \State Prompt $M_i$ and $M_j$ to choose a horse (with option to use system-generated one)
                \State Log choices and update game state
            \EndWhile
            \State Record win/loss outcome of this tournament
            \State Compute log Bayes factor for $M_i$ and $M_j$'s action sequences \Comment{See \autoref{sec:bayesfactor} for definition}
        \EndFor
        \State Average log Bayes factors over $K$ tournaments
        \State Store results in $W$ and $B$
    \EndFor
    \State \Return Win/loss matrix $W$ and Bayes factor double matrix $B$
\end{algorithmic}
\end{algorithm}
\FloatBarrier

\paragraph{Prompt Variants}

We evaluate three types of prompts to assess how framing affects strategic behavior:

\begin{enumerate}
    \item \textbf{Framed Prompt}: Mentions the historical context (Tian Ji), suggests devising clever strategies to "win smartly".
    \item \textbf{Neutral Prompt}: Abstract description of the game rules, no game-theoretic hints.
    \item \textbf{Hinted Prompt}: Explicitly states that uniform random selection over remaining horses is the Nash equilibrium.
\end{enumerate}

These prompt variants allow us to study whether models can internalize theoretical insights or adapt strategically through repeated interactions.

\subsection{Metrics and Analysis}

We measure two primary aspects of LLM behavior:
\begin{itemize}
    \item \textbf{Competitive Performance}: Who wins more tournaments?
    \item \textbf{Degree of Randomization}: How likely is the model to adopt the system-provided random option?
\end{itemize}

\paragraph{Win/Loss Matrix}
We calculate the net relative wins between two LLMs over $ K = 10 $ tournaments. The result is presented as a 5x5 matrix, where each entry $ (i, j) $ represents the number of times model $ j $ outperforms model $ i $, minus the reverse. Positive values indicate dominance of the \textbf{column} model over the \textbf{row} model, while negative values suggest the opposite.

\paragraph{Bayes Factor Computation} \label{sec:bayesfactor}

To quantitatively assess the degree to which LLMs follow a truly randomized strategy, we adopt a Bayesian perspective based on likelihood comparison between two hypotheses:

\begin{itemize}
    \item $ \mathcal{H}_1 $: \textbf{Randomized Strategy} \textemdash The agent chooses to randomize and tends to use the system-provided random option.
    \item $ \mathcal{H}_2 $: \textbf{Strategic Deviation} \textemdash The agent adheres to its own internal strategy, disregarding the system-provided random option entirely; in this sense, its choices exhibit a form of relative randomness with respect to the reference provided by the system's suggestion.
\end{itemize}

For each tournament, we compute the \textbf{log Bayes factor}:

\[
\mathcal{B}_{12} = \log \frac{\mathbb{P}\left[\text{data} \mid \mathcal{H}_1\right]}{\mathbb{P}\left[\text{data} \mid \mathcal{H}_2\right]} = \sum_{i=1}^{N} \log \mathbb{P}\left[\text{choice}_i \mid \mathcal{H}_1\right] - \sum_{i=1}^{N} \log \mathbb{P}\left[\text{choice}_i \mid \mathcal{H}_2\right]
\]

Where:
\begin{itemize}
    \item $ \mathbb{P}[\text{choice}_i \mid \mathcal{H}_1] = 
  \begin{cases}
    1 - \frac{|\text{legal choices}| - 1}{|\text{legal choices}|}\eta & \text{if choice}_i \text{ matches the system-generated random choice} \\
    \frac{\eta}{|\text{legal choices}|} & \text{otherwise}
  \end{cases} $
  \item $ \mathbb{P}[\text{choice}_i \mid \mathcal{H}_2] = \frac{1}{|\text{legal choices}|} $ uniformly across all available options.
\end{itemize}
Here, $ \eta = 0.4 $ is a conservative estimate of the model's baseline tendency to abandon the random suggestion when it intends to randomize. The larger the value of $ \eta $, the greater the model's tendency to deviate from the system-generated random choices and follow its own strategy.

This formulation allows us to measure each agent’s behavior along a spectrum—from fully randomized ($ \mathcal{H}_1 $ favored) to highly deterministic ($ \mathcal{H}_2 $ favored)—and observe how this tendency evolves through repeated tournaments and prompt variations.

\section{Experimental Setup}

\paragraph{Model Selection}
We evaluate five representative large language models (LLMs) across different architectures and capabilities. These include:  
\begin{itemize}
    \item Weak models
    \begin{itemize}
        \item \verb|doubao-1-5-pro-256k-250115|
        \item \verb|gpt-4.1-mini-2025-04-14|
    \end{itemize}
    \item Strong models
    \begin{itemize}
        \item \verb|gpt-4.1-2025-04-14|
        \item \verb|o4-mini|
        \item \verb|gemini-2.5-flash-preview-04-17(no thinking)|
    \end{itemize}
\end{itemize}

These models are selected to cover a range of proprietary systems, enabling us to compare performance across model families.

\paragraph{Play Configurations}
Each model is pitted against every other model, including itself, in a round-robin format. For each pair of models, we conduct $ K = 10 $  consecutive tournaments to allow for potential adaptation based on prior interactions. Each tournament involves $ N = 7 $ horses per player, resulting in a total of 7 rounds per tournament.

\paragraph{Implementation Details}
At the beginning of each round, the system provides a system-generated random choice sampled uniformly from the set of remaining horses for each player. This random choice is private to each model and not exposed to its opponent. The model is instructed that it may use this random choice to guide its action if it wishes to randomize.

The temperature parameter of all models is kept at default settings, ensuring no artificial inflation or suppression of stochasticity. No external memory or fine-tuning is used — models interact with the game solely through the prompt interface.

All experiments are conducted with fixed random seeds to ensure reproducibility of the system-generated choices.

\paragraph{Ablation Studies}
We perform ablation studies across three prompt conditions:
\begin{enumerate}
    \item \textbf{Framed Prompt}: Emphasizes narrative elements (e.g., Tian Ji, horse racing strategy), encouraging heuristic or deterministic strategies.
    \item \textbf{Neutral Prompt}: Abstract description of the rules without any strategic framing.
    \item \textbf{Hinted Prompt}: Explicitly informs the model that uniform randomization over remaining actions constitutes a Nash equilibrium.
\end{enumerate}

We analyze how these prompts affect the models' ability to randomize and adapt strategically, both when playing against stronger and weaker models.

\section{Results and Analysis}
\subsection{Win/Loss Outcomes}
\begin{figure}
  \centering
  \begin{minipage}[b]{0.45\textwidth}
    \centering
    \fbox{\includegraphics[width=\textwidth]{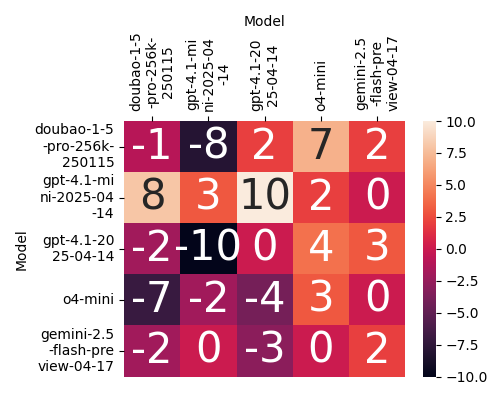}}
  \end{minipage}
  \hfill
  \begin{minipage}[b]{0.45\textwidth}
    \centering
    \fbox{\includegraphics[width=\textwidth]{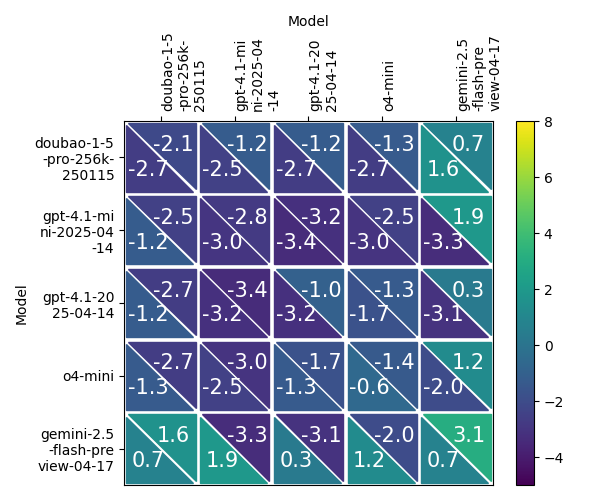}}
  \end{minipage}
  \caption{Win/lose matrix and average Bayes factor of each pair of competing models under framed prompt}
  \label{fig:figure1}
\end{figure}

Across all experiments, more capable models consistently outperform less capable ones. This is evident in the win/loss matrices for each prompt type (Framed, Neutral, Hinted), as shown in \autoref{fig:figure1}. The matrix on the left represent the net relative wins between two LLMs over $ K=10 $ tournaments. Each entry $(i, j)$ in these matrices represents the number of times model $j$ outperforms model $i$, minus the reverse. Positive values indicate dominance of the column model over the row model, while negative values suggest the opposite.

\begin{figure}
  \centering
  \begin{minipage}[b]{0.45\textwidth}
    \centering
    \includegraphics[width=\textwidth]{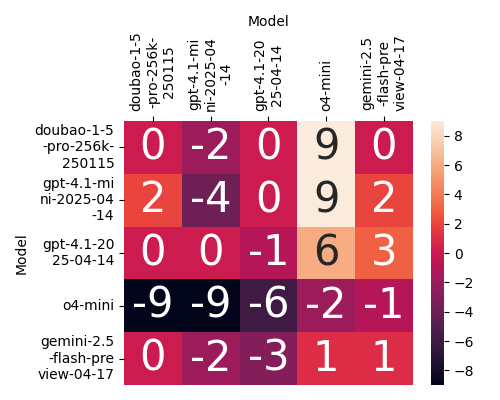}
  \end{minipage}
  \hfill
  \begin{minipage}[b]{0.45\textwidth}
    \centering
    \includegraphics[width=\textwidth]{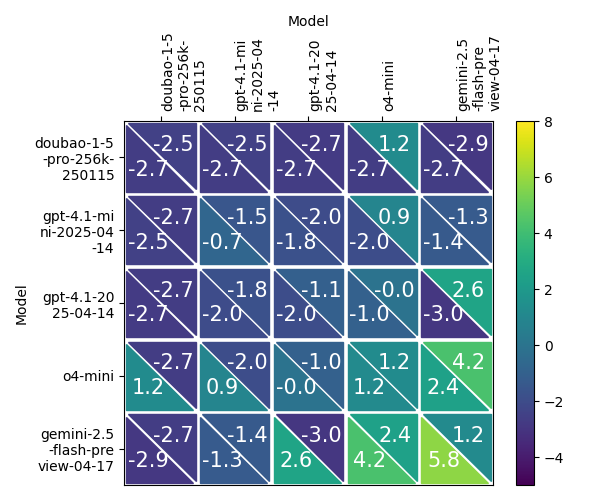}
  \end{minipage}
  \caption{Win/lose matrix and average Bayes factor of each pair of competing models under neutral prompt}
  \label{fig:figure2}
\end{figure}

\begin{figure}
  \centering
  \begin{minipage}[b]{0.45\textwidth}
    \centering
    \includegraphics[width=\textwidth]{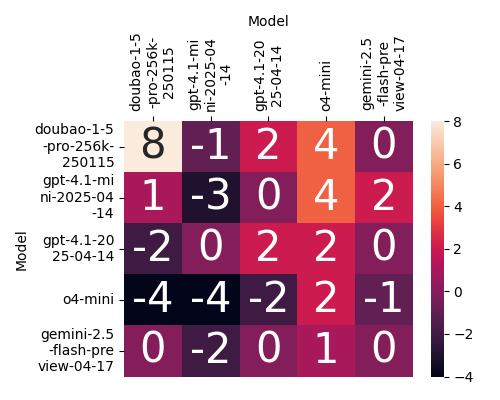}
  \end{minipage}
  \hfill
  \begin{minipage}[b]{0.45\textwidth}
    \centering
    \includegraphics[width=\textwidth]{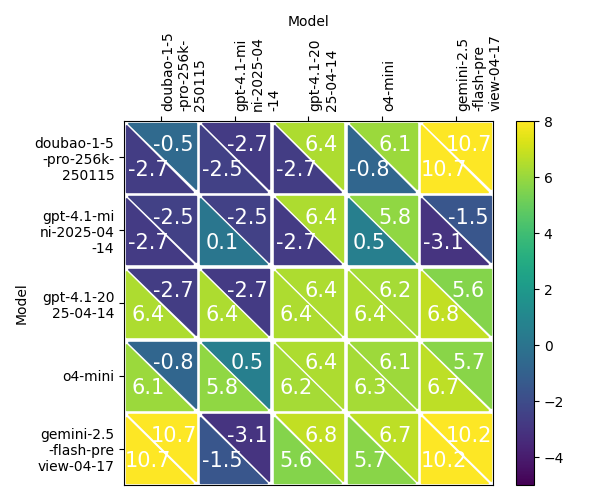}
  \end{minipage}
  \caption{Win/lose matrix and average Bayes factor of each pair of competing models under hinted prompt}
  \label{fig:figure3}
\end{figure}

\subsection{Bayes Factor Analysis}
Less capable models show low Bayes factors regardless of prompt, indicating deterministic behavior. This is evident in the log Bayes factor matrices for all prompt types. Models like \verb|doubao-1-5-pro-256k-250115| consistently exhibit low log Bayes factors, suggesting they rely heavily on deterministic strategies.

Stronger models exhibit high Bayes factors under hinted prompts, confirming deliberate randomization. As seen in \autoref{fig:figure3}, strong models have high log Bayes factors when paired with other strong models under hinted prompts, indicating their williness to follow the system's random numbers.

Under framed prompts, even strong models sometimes fall into deterministic traps, mimicking suboptimal human behavior. This is observed in \autoref{fig:figure1}, where strong models occasionally show lower log Bayes factors, suggesting they may default to deterministic heuristics despite hints.

\subsection{Case Studies}
Qualitative analysis of model responses across different prompts reveals a diverse range of strategic behaviors. For instance, the model \verb|doubao-1-5-pro-256k-250115| consistently employed heuristic strategies—such as always deploying its fastest horse early—and failed to adapt or learn from repeated gameplay, regardless of prompt type. This rigid behavior highlights a lack of strategic flexibility and an inability to recognize the value of randomization.

Conversely, under the framed prompt, \verb|o4-mini| initially hypothesized that the maximal entropic strategy corresponds to the Nash equilibrium. However, in subsequent rounds, it became uncertain and speculated that the optimal strategy might be more complex than uniform randomization. This case illustrates how even models capable of initial reasoning can be misled by contextual framing, echoing human cognitive biases.

These examples, supported by screenshots of key turns (see Appendix), demonstrate both alignment and divergence from theoretically optimal play, offering insight into how LLMs interpret and respond to strategic uncertainty under different prompting conditions.

\subsection{Limitations}
The current experimental setup assumes $N=7$, full observability, and symmetric information between players. While these simplifications facilitate controlled evaluation, they limit the generalizability of findings to real-world scenarios involving asymmetric information, partial observability, or variable player roles.

Moreover, LLMs lack mechanisms for reinforcement learning during gameplay—unlike humans, who may adjust strategies based on experiential feedback, LLMs rely solely on prompt context and pre-trained knowledge. This constraint affects their ability to iteratively refine strategies over time.

Future work should explore variations in game parameters, such as increasing or decreasing $N$, introducing more players, and incorporating imperfect information settings. Additionally, integrating reinforcement learning frameworks could help train models to better recognize and adopt randomized strategies through experience.

\section{Conclusion}
This work investigates whether large language models (LLMs) can recognize situations where randomization is strategically optimal and whether they can deliberately choose to randomize. By designing a novel, structurally complex game inspired by the Tian Ji Horse Race—where the Nash equilibrium corresponds to a maximal entropy strategy—we cleanly single out the decision to randomize from the act of generating randomness, enabling precise evaluation. Our experimental framework evaluates five LLMs across various prompt conditions, revealing that:
\begin{itemize}
    \item LLMs exhibit varying degrees of strategic randomization depending on model strength and prompt design.
    \item Weaker models tend to behave deterministically regardless of prompts.
    \item Stronger models can be prompted toward more randomized behavior, especially when explicitly told that uniform random play constitutes equilibrium.
    \item When facing weaker opponents, even strong models abandon randomization and exploit deterministic patterns.
\end{itemize}

These findings suggest that while some LLMs are capable of reasoning about randomization as a strategy, this ability is neither consistent nor robust. Future efforts should focus on improving both architectural and training mechanisms that encourage abstract strategic reasoning and learning from repeated interactions. This work provides a principled foundation for evaluating and advancing LLMs in domains where strategic uncertainty and optimal randomization are essential.

\bibliographystyle{plainnat} % Style compatible with natbib
\bibliography{references} % Assumes your .bib file is named 'references.bib'

\section{Appendix}
\paragraph{Contributions}

This work was entirely designed, implemented, analyzed, and written by the sole author.

\paragraph{Computational Resources and Cost}

The experiments were conducted using API access to multiple large language models provided by \verb|Meituan|, under an academic sponsorship arranged by instructor \verb|ying.wen@sjtu.edu.cn|. No financial cost was incurred.

\paragraph{Ethical Considerations}

No personal or sensitive data was collected during the course of this study. All interactions were synthetic and model-generated.

%%%%%%%%%%%%%%%%%%%%%%%%%%%%%%%%%%%%%%%%%%%%%%%%%%%%%%%%%%%%

\end{document}